\documentclass{article}
\usepackage{spconf,amsmath,graphicx}
\usepackage{amssymb,amsthm}
\usepackage{pst-all,psfrag}
\newtheorem{theorem}{Theorem}[section]

\newtheorem{lemma}[theorem]{Lemma}

\usepackage{framed}

\newcommand{\transp}[1]{{#1}^T} 

\newcommand{\condmiyandlgiveni}{I(\mathbf{Y};\dicidx| \mathbf{\suppidx})}

\DeclareMathOperator*{\argmin}{argmin}

\newcommand\coefflen{p}
\newcommand{\measlen}{m} 
\newcommand{\samplesize}{N}
\newcommand{\sparsity}{s}
\newcommand{\obsidx}{k}
\newcommand{\suppidx}{i}
\newcommand{\dicidx}{l}
\newcommand{\rmv}{\hspace*{-.2mm}}
\newcommand\defeq{:=}
\DeclareMathOperator*{\trace}{Tr}

\newcommand{\condcovyi}{\mathbf{\Sigma}_{y|\suppidx}}
\newcommand{\condcovyik}{\mathbf{\Sigma}_{y|\suppidx_{\obsidx}}}
\def \expect {{\rm E} }
\def \prob {{\rm P} }
\DeclareMathOperator{\supp}{supp}

\newcommand{\minimaxrisk}{\varepsilon}

\allowdisplaybreaks

\title{Performance Limits of Dictionary Learning for Sparse Coding}
\name{Alexander Jung$\rmv^{a}\rmv$, Yonina C. Eldar$^{b}\rmv\rmv$, Norbert G\"{o}rtz$^{a}\rmv\rmv$}

\address{\normalsize $^a$Institute of Telecommunications, Vienna University of Technology, Austria; \{ajung, norbert.goertz\}@nt.tuwien.ac.at\\[1mm]
\normalsize $^b$Technion---Israel Institute of Technology, Israel; e-mail: yonina@ee.technion.ac.il }

\begin{document}
\maketitle

\begin{abstract}
We consider the problem of dictionary learning under the assumption that the observed signals can be represented as sparse linear combinations of the columns of 
a single large dictionary matrix. In particular, we analyze the minimax risk of the dictionary learning problem which governs the mean squared error (MSE) performance of any learning scheme, regardless of its computational complexity. 
By following an established information-theoretic method based on Fano's inequality, we derive a lower bound on the minimax risk for a given dictionary learning problem. 
This lower bound yields a characterization of the sample-complexity, i.e., a lower bound on the required number of observations such that consistent dictionary learning schemes exist. 
Our bounds may be compared with the performance of a given learning scheme, allowing to characterize how far the method is from optimal performance. 
\end{abstract}

\section{Introduction}
\label{sec_intro} 

\vspace{-.2mm}

Consider observing $\samplesize$ signals $\mathbf{y}_{k} \in \mathbb{R}^{\measlen}$, $k=1,\ldots,\samplesize$, which are assumed to be sparse linear combinations of the 
columns of an underlying dictionary $\mathbf{D} \in \mathbb{R}^{\measlen \times \coefflen}$.
Each signal $\mathbf{y}_{k}$ is an i.i.d.\ realization of the random vector 
\vspace*{-2mm}
\begin{equation}
\label{equ_linear_model}
\mathbf{y} = \mathbf{D} \mathbf{x} + \mathbf{w}. 
\vspace*{-2mm}
\end{equation} 
The matrix $\mathbf{D}Ê\in \mathbb{R}^{\measlen \times \coefflen}$, with $\coefflen \geq \measlen$, is the underlying dictionary we wish to learn. The random vector $\mathbf{x} \in \mathbb{R}^{\coefflen}$ is a sparse coefficient vector and $\mathbf{w} \in \mathbb{R}^{\measlen}$ 
denotes zero-mean additive white Gaussian noise with variance $\sigma^{2}>0$, i.e., $\mathbf{w} \sim \mathcal{N}(\mathbf{0},\sigma^{2} \mathbf{I})$. The dictionary learning problem is relevant to a wide range of applications and has been studied extensively. In particular, dictionary learning is applied to \emph{Big Data} applications aiming at discovering an intrinsic low dimensional structure in very high-dimensional data, in order to make 
the processing of this data flood tractable. 

\emph{State of the Art:}
A variety of (locally) efficient learning schemes have been proposed and analyzed in the literature (e.g., \cite{Schnass2014,KSVD,Jenatton2012,GribonvalSchnass2010,YaghDicLearn2009,BilinAMP}). 
In \cite{BilinAMP} the authors apply a variant of the \emph{approximate message passing} scheme \cite{AMP2009PNAS} to the dictionary learning problem. 
The works in \cite{KSVD,Jenatton2012,GribonvalSchnass2010,YaghDicLearn2009,Mairal2009ICML} consider 
estimates of the dictionary obtained by solving the (non-convex) minimization problem
\begin{equation}
\label{equ_minimization_ell_1_dic_learning}
\min_{\mathbf{D},\mathbf{X}}  \| \mathbf{Y} - \mathbf{D} \mathbf{X} \|_{\text{F}}^{2} + \lambda \|\mathbf{X}\|_{1},
\end{equation}
where the $k$th columns of $\mathbf{X}$ and $\mathbf{Y}$ are given by the $k$th i.i.d.\ realizations $\mathbf{y}_{k}$ and $\mathbf{x}_{k}$, respectively, and $\|\mathbf{X}\|_{1} \defeq \sum_{k,l} |X_{k,l}|$. 
The authors of \cite{KSVD,Jenatton2012,GribonvalSchnass2010} give upper bounds on the distance between the generating dictionary and the nearest local minimum of \eqref{equ_minimization_ell_1_dic_learning}. Based on these 
characterizations of the local minima, it has been shown in \cite{GribonvalSchnass2010}, for the noiseless and square dictionary setting, that $\samplesize \propto \coefflen \log( \coefflen)$ observations are sufficient to 
guarantee local identifiability of the generating dictionary. By contrast, the authors of \cite{Jenatton2012} obtain a sample-complexity of 
$\samplesize \propto \coefflen^{3} \measlen$ in the case of overcomplete dictionaries and noisy observations. 
The analysis presented in \cite{KSVD,Jenatton2012,GribonvalSchnass2010} is conceptually different from our analysis, since we focus on the (worst-case) MSE of learning schemes, whereas \cite{KSVD,Jenatton2012,GribonvalSchnass2010} characterize the existence of local minima (of \eqref{equ_minimization_ell_1_dic_learning}) close to the generating dictionary. 
We would also like to mention an exciting recent line of work \cite{spwawr12,argemo13,aganne13,aganjaneta13} presenting dictionary learning schemes that are proven to globally 
recover the generating dictionary.


\emph{Contribution:}
By now there have been proposed quite a few dictionary learning schemes, whose performance is theoretically analyzed in terms of a characterization of the sample size sufficient for (local) identification of the generating dictionary. 
However, an investigation of fundamental performance limits for the dictionary learning problem seems to be missing. Here, we close this gap and present a lower bound on the minimax risk for the dictionary learning problem, where the estimation quality is measured by the Frobenius norm. This bound applies to any algorithm, regardless of its computational complexity and seems to be the first analysis that targets directly the MSE of learning schemes. 
For the derivation of the lower bound, we make use of an established information-theoretic approach to 
minimax estimation, which is based on Fano's inequality \cite{coverthomas}. 
Although this approach has been successfully applied to several other (sparse) estimation problems \cite{WangWain2010,Wain2009TIT,CandesDavenport2013,CaiZhouSparseCov}, the adaptation  
of this method to the problem of dictionary learning for sparse coding seems to be new. 

\emph{Outline of the Paper:}
We begin in Section \ref{SecProblemFormulation} with a formalization of the problem setup and discuss the adaption of the information-theoretic proof method (for lower bounding the minimax risk) to this setting. 
A lower bound on the minimax risk for dictionary learning is presented in Section \ref{sec_main_result}. A sketch of the proof is given in Section \ref{proof_architecture_main_result}. 

\emph{Notation:}
Given a natural number $k \in \mathbb{N}$, we define the set $[k] \triangleq \{1,\ldots,k\}$. For a matrix $\mathbf{A} \in \mathbb{R}^{\measlen \times \coefflen}$, we denote its Frobenius norm by 
$\| \mathbf{A} \|_{\text{F}} \triangleq \sqrt{\trace \{\mathbf{A} \mathbf{A}^{T}Ê\}}$. The $k$th column of the identity matrix is denoted by $\mathbf{e}_{k}$. 
The complementary Kronecker delta is denoted by $\bar{\delta}_{l,l'}$, where $\bar{\delta}_{l,l'} = 0$ if $l=l'$ and is equal to one otherwise. 
The determinant of a square matrix $\mathbf{C}$ is denoted by $|Ê\mathbf{C}|$. We denote by $\expect_{\mathbf{Z}}\{\cdot\}$ the expectation w.r.t. the distribution 
of the random vector or matrix $\mathbf{Z}$.

\vspace{-.7mm}

\section{Problem Formulation}
\label{SecProblemFormulation}
\subsection{The Dictionary Learning Problem}
Consider the model \eqref{equ_linear_model}. We collect the measurements into the observation matrix  
\vspace*{-2mm}
\begin{equation}
\mathbf{Y} \defeq \begin{pmatrix} \mathbf{y}_{1},\ldots,\mathbf{y}_{\samplesize} \end{pmatrix} \in \mathbb{R}^{\measlen \times \samplesize},
\vspace*{-2mm}
\end{equation}
where $\mathbf{y}_{k}$ is an i.i.d. realization of the random vector given by \eqref{equ_linear_model}. 
The underlying generating dictionary $\mathbf{D}$ is modeled as deterministic but unknown. 
We assume the columns of $\mathbf{D}$ to be normalized, i.e., 
\vspace*{-2mm}
\begin{equation}
\label{equ_def_dictionary_set_normalized_cols}
\mathbf{D} \in \mathcal{D} \triangleq \{ \mathbf{B} \in \mathbb{R}^{\measlen \times \coefflen} | \| \mathbf{B} \mathbf{e}_{j} \|_{2} = 1 \mbox{, for all } j \in [\coefflen] \}. 
\vspace*{-1mm}
\end{equation}
The set $\mathcal{D}$ is known as the \emph{oblique manifold} \cite{Jenatton2012}. Moreover, we assume the true dictionary $\mathbf{D}$ 
to be obtained as a small perturbation of a known ``reference dictionary'' $\mathbf{D}_{0}$. In particular, for some small radius $r>0$, we require 
\vspace*{-2mm}
\begin{equation} 
\mathbf{D} \in \mathcal{X}(\mathbf{D}_{0},r) \defeq \{ \mathbf{D}' \in \mathcal{D}: \| \mathbf{D} - \mathbf{D}_{0} \|_{\rm{F}} \leq r \}
\label{equ_def_par_set_small_perturb} 
\vspace*{-1mm}
\end{equation}

The statistics of the coefficient vector $\mathbf{x}$ is modeled such that it is a strictly $\sparsity$-sparse vector. In particular, we introduce the random variable $\suppidx$, 
which is chosen uniformly at random (u.a.r.) from the set $[\binom{\coefflen}{\sparsity}]$. A specific value of $\suppidx$ represents 
a certain index set $\mathcal{S}(\suppidx) \subseteq [\coefflen]$ containing $\sparsity$ different indices. More formally, the map 
\begin{equation}
\label{equ_index_mapping_subsets}
\mathcal{S}(\cdot): \bigg[ \binom{\coefflen}{\sparsity} \bigg] \rightarrow \mathcal{E} \triangleq \{ \mathcal{I} \subseteq [\coefflen], | \mathcal{I} | = \sparsity \}
\end{equation}
is a bijection from the first $\binom{\coefflen}{\sparsity}$ natural numbers to the set $\mathcal{E}$ of all size-$\sparsity$ subsets $\mathcal{I}$ of $[\coefflen]$. 

The random variable $\suppidx$ selects the active coefficients of $\mathbf{x}$, i.e., 
\begin{equation}
\label{equ_representation_random_masking_z_for_x}
\supp ( \mathbf{x} ) = \mathcal{S}(\suppidx) \mbox{, and } \mathbf{x}_{\mathcal{S}(\suppidx)} \sim \mathcal{N}(\mathbf{0}, \sigma_{a}^{2} \mathbf{I}). 
\end{equation}
The (unconditional) covariance matrix of the sparse coefficient vector $\mathbf{x}$ is given by 
\begin{equation}
\label{equ_expr_cov_matrix_x}
\mathbf{\Sigma}_{x} \triangleq \expect \{ \mathbf{x} \transp{\mathbf{x}} \}  = (\sparsity/\coefflen) \sigma_{a}^{2} \mathbf{I}.
\end{equation}
We define the signal to noise ratio of the observation model \eqref{equ_linear_model} as $\mbox{SNR} \defeq (\sigma_{a}/\sigma)^{2}$.

Since the columns of $\mathbf{Y}$ are i.i.d. realizations of the vector $\mathbf{y}$ in \eqref{equ_linear_model}, 
the conditional probability density function (pdf) of the observation $\mathbf{Y}$, given the $\samplesize$ i.i.d.\ realizations $\mathbf{\suppidx} = (\suppidx_{1},\ldots,\suppidx_{\samplesize})$ of the random support index $\suppidx$, is 
\begin{equation}
\label{equ_expr_pdf_y_given_i}
f_{\mathbf{D}}(\mathbf{Y}|\mathbf{\suppidx}) = \prod_{\obsidx \in [\samplesize]} \frac{ \exp \big( - (1/2) \mathbf{y}^{T}_{\obsidx}  \condcovyik^{-1} \mathbf{y}_{\obsidx} \big)}{(2 \pi)^{\measlen/2} \big| \condcovyik \big|^{1/2} }. \nonumber
\end{equation} 
Here, $\condcovyi \triangleq \expect \big\{ \mathbf{y} \transp{\mathbf{y}} \big| \suppidx \big\}$ denotes the conditional covariance matrix  of $\mathbf{y}$, given $\suppidx$, and 
reads explicitly as  
$\condcovyi \!=\! \sigma_{a}^{2} \mathbf{D}_{\mathcal{S}(\suppidx)}  \mathbf{D}^{T}_{\mathcal{S}(\suppidx)}+ \sigma^{2} \mathbf{I}$. 

We note that any learning scheme based on the model \eqref{equ_linear_model} faces an intrinsic sign and permutation ambiguity for the dictionary $\mathbf{D}$. Indeed, by observing $\mathbf{Y}$ only, one cannot  
distinguish between dictionaries which are related via column permutations and sign-flips of the columns \cite{Jenatton2012,GribonvalSchnass2010}. 
While we do not take this intrinsic ambiguity into account explicitly, our results are meaningful as they apply to dictionary learning problems where the 
true dictionary belongs to the (small) neighborhood $\mathcal{X}(\mathbf{D}_{0},r)$ of a known reference dictionary $\mathbf{D}_{0}$. 

We investigate the fundamental limits on the accuracy achievable by any learning scheme producing an estimate $\widehat{\mathbf{D}}(\mathbf{Y})$ of the underlying dictionary based on the observation $\mathbf{Y}$. For the moment, suppose that we have access to the coefficients $\mathbf{x}$ in \eqref{equ_linear_model} and the estimate $\widehat{\mathbf{D}}$ is held fixed, i.e., does not depend on the observation $\mathbf{Y}$. Then, we obtain for the prediction error, when using the estimate $\widehat{\mathbf{D}}$ instead of the generating dictionary $\mathbf{D}$,  
\vspace*{-2mm}
\begin{align} 
\label{equ_relation_prediction_error_Frob_norm}
\expect_{\mathbf{x}} \{\| \mathbf{D} \mathbf{x} - \widehat{\mathbf{D}} \mathbf{x}  \big\|^{2}\} 
& \stackrel{\eqref{equ_expr_cov_matrix_x}}{=} 
(\sparsity/\coefflen)\sigma_{a}^{2} \| \mathbf{D} - \widehat{\mathbf{D}} \|_{\rm{F}}^{2}.Ê
\end{align} 
Therefore, the prediction error is proportional to the squared Frobenius norm of the estimation error $\mathbf{D}-\widehat{\mathbf{D}}$. 
Based on \eqref{equ_relation_prediction_error_Frob_norm}, we measure the accuracy of a specific learning scheme $\widehat{\mathbf{D}}(\cdot)$ by the MSE 
$\varepsilon(\mathbf{D},\widehat{\mathbf{D}}(\cdot)) \triangleq \expect _{\mathbf{Y}}\{ \| \widehat{\mathbf{D}}(\mathbf{Y}) - \mathbf{D} \|^{2}_{\text{F}} \}$. 
Note that the MSE depends on the underlying generating dictionary $\mathbf{D}$ and the learning scheme $\widehat{\mathbf{D}}(\cdot)$. 

Define the minimax risk $\minimaxrisk$ for the problem of learning the dictionary $\mathbf{D}$ based on the observation of $N$ i.i.d. realizations of $\mathbf{y}$ in \eqref{equ_linear_model}, as
\begin{equation}
\label{equ_def_minimax_problem}
\minimaxrisk \triangleq \inf_{\widehat{\mathbf{D}}} \sup_{\mathbf{D} \in \mathcal{X}(\mathbf{D}_{0},r)}  \varepsilon(\mathbf{D},\widehat{\mathbf{D}}(\cdot)). 
\end{equation} 
The minimax risk $\varepsilon$ will in general depend on the number of observations $\samplesize$, the dimension $\measlen$ of the observed signals, the 
number of signal expansion coefficients $\coefflen$, the sparsity degree $\sparsity$ and the variance parameters $\sigma_{a}^{2}$ and $\sigma^{2}$. 
However, to lighten notation, we will not make this dependence explicit. Our goal is to develop a lower bound on $\varepsilon$ using an information-theoretic method. 

Having a lower bound for the minimax risk allows us to asses the performance of a given dictionary learning scheme. In particular, 
if the MSE of a given algorithm is close to the minimax risk, or a lower bound to it, then there is little point to hope for finding improved techniques with substantially better 
performance. 

\subsection{Information Theory of Dictionary Learning}

Our approach to bounding the minimax risk $\varepsilon$ of \eqref{equ_def_minimax_problem} is to use the information-theoretic method put forward in \cite{BinYuFestschrift1997,WangWain2010,CandesDavenport2013}. 
However, the key challenge in applying this technique is the fact that the vector $\mathbf{y}$ given by \eqref{equ_linear_model} does not follow a multivariate normal distribution. Indeed, due to the prior model for the coefficient vector $\mathbf{x}$ (cf.\ \eqref{equ_representation_random_masking_z_for_x}), the 
vector $\mathbf{y}$ follows a Gaussian mixture model, with a mixture component associated with each specific value of the support index $\suppidx$. 

In order to apply the information-theoretic technique, it is 
necessary to have a precise characterization of the mutual information $I(\mathbf{Y}; \dicidx)$ between the observation $\mathbf{Y}$ and a random index $\dicidx$ which 
selects the generating dictionary $\mathbf{D} = \mathbf{D}^{(\dicidx)}$ u.a.r. from a finite set $\mathcal{D}_{0} \subseteq \mathcal{D}$. 
Obtaining a bound on $I(\mathbf{Y};\dicidx)$ typically involves the analysis of the Kullback Leibler (KL) divergence between 
the distributions of $\mathbf{Y}$ implied by different dictionaries $\mathbf{D}=\mathbf{D}^{(\dicidx)}$. 
However, exact characterizations of the KL divergence between Gaussian mixture models is in general not possible and one has to 
resort to approximations or bounds \cite{HersheyOlsen2007}. 

A main conceptual contribution of this work is a strategy to avoid evaluating 
KL divergences between Gaussian mixture models. Instead, we rely on the following decomposition, which follows from the chain rule for mutual information, 
\vspace*{-2mm}
\begin{align}
\label{equ_decomp_mutual_inf_chain_rule}
I(\mathbf{Y}; \dicidx) & = I(\mathbf{Y},\mathbf{\suppidx};\dicidx) - I( \dicidx ; \mathbf{\suppidx} | \mathbf{Y}) \nonumber \\[1mm]
    & = \condmiyandlgiveni + \underbrace{I(\dicidx;\mathbf{\suppidx})}_{=0}   - I( \dicidx ; \mathbf{\suppidx} | \mathbf{Y})  \nonumber \\ 
    & = \condmiyandlgiveni -  I( \dicidx ; \mathbf{\suppidx} | \mathbf{Y}). \\[-8mm] 
    \nonumber 
\vspace*{-1mm}
\end{align}
Here, $\condmiyandlgiveni$ denotes the conditional mutual information between the observation $\mathbf{Y}$ and the random index $\dicidx$, given 
the support indices $\mathbf{\suppidx} = (\suppidx_{1},\ldots,\suppidx_{\samplesize})$. The components of the decomposition in \eqref{equ_decomp_mutual_inf_chain_rule} have particular interpretations. The term $\condmiyandlgiveni$ characterizes 
the difficulty of detecting the (index of the) generating dictionary $\mathbf{D} = \mathbf{D}^{(\dicidx)}$, if we had access to the indices $\suppidx_{k}$ selecting the active coefficients of $\mathbf{x}_{k}$. 
The second term, i.e., $I(\dicidx ; \mathbf{\suppidx} | \mathbf{Y})$ quantifies the dependence between the support of the sparse coefficient vector $\mathbf{x}$ and the (index $\dicidx$ of the) generating dictionary $\mathbf{D} = \mathbf{D}^{(\dicidx)}$, after observing $\mathbf{Y}$. 

Since $I( \dicidx ; \mathbf{\suppidx} | \mathbf{Y})\geq 0$ \cite[Ch. 2]{coverthomas}, we can upper bound $I(\mathbf{Y}; \dicidx)$ by upper bounding $\condmiyandlgiveni$. 
Note that, conditioned on the support index $\suppidx$, the data vector $\mathbf{y}$ in \eqref{equ_linear_model} follows a normal distribution with covariance matrix $\condcovyi$, which renders 
the problem of upper bounding $\condmiyandlgiveni$ tractable. We detail this proof technique in Section \ref{proof_architecture_main_result}. 
\vspace*{-2mm}


\section{A Lower Bound on the Minimax Risk} 
\label{sec_main_result}

A typical requirement for sparse (compressed sensing) recovery to work well, even when the dictionary $\mathbf{D}$ in \eqref{equ_linear_model} is known, is the 
validity of \cite{RauhutFoucartCS,EldarKutyniokCS}
\vspace*{-3mm}
\begin{equation}
\label{equ_condition_m_geq_s_log_n}
m \geq c_{0} \sparsity \log( \coefflen/\sparsity ), 
\vspace*{-2mm}
\end{equation}
with some absolute constant $c_{0}$.
Since we consider the more difficult problem of dictionary learning, i.e., we treat the dictionary as an unknown parameter, we expect \eqref{equ_condition_m_geq_s_log_n}
to be a necessary requirement for the existence of accurate dictionary learning schemes. 

Our main result is the following lower bound on the minimax risk for a given dictionary learning problem.
\begin{theorem}
\label{thm_main_result}
Consider a dictionary learning problem based on $\samplesize$ i.i.d. observations following the model \eqref{equ_linear_model} and the 
true dictionary satisfying \eqref{equ_def_par_set_small_perturb} with $r\leq1/\sqrt{\coefflen}$. Then, if 
\vspace*{-2mm}
\begin{equation} 
\label{equ_cond_coefflen_measlen}
\coefflen >  64 \mbox{,  and }  \measlen \geq 192 \sparsity (9 + 2\log(\coefflen/ \sparsity) ),
\vspace*{-2mm}
\end{equation}
the minimax risk $\minimaxrisk$ is lower bounded as
\vspace*{-2mm}
\begin{equation}
\label{equ_conditions_s_2_first_charac_minimax_risk}
\minimaxrisk \geq  \min \bigg\{ r^{2}/16,  \frac{\emph{SNR}^{-1} \coefflen^{2}}{5120  \samplesize \sparsity}  \bigg\}.
\vspace*{-2mm}
\end{equation}
\end{theorem}

We highlight the fact that Theorem \ref{thm_main_result} does not place any assumptions (like incoherence or restricted isometry properties) on the underlying generating dictionary.

For sufficiently large sample-size $\samplesize$, such that $\samplesize \!\gg\! \coefflen^{2}/\sparsity$, the second bound in \eqref{equ_conditions_s_2_first_charac_minimax_risk} will be in force. 
This bound shows a dependence on the sample-size via $1/N$ which clearly makes sense. Indeed, 
by averaging the outcomes of a learning scheme over blocks of independent observations the MSE is expected to scale inversely proportional to the sample size $\samplesize$. 
This dependence of the MSE on the sample-size is also observed in the empirical results of simulation studies for specific learning schemes in \cite{Jenatton2012,Schnass2014}. Moreover, the theoretic results presented in \cite{Jenatton2012,SchnassIKTM2014} suggest that the estimation error of certain learning schemes, measured by the squared Frobenius norm, scales inversely proportional to $\samplesize$.

For the case of constant sparsity, i.e., when $\sparsity \leq C_{0}$ for some constant (independent of $\coefflen$) our lower bound 
scales as $\Theta(\coefflen^{2}/N)$, suggesting a sample-complexity of $\Theta(p^{2})$. This scaling is considerably smaller than the sample complexity $\mathcal{O}(\coefflen^3 \measlen)$, which \cite{Jenatton2012} proved to be sufficient in the noisy and over-complete case, such that the estimator based on minimizing \eqref{equ_minimization_ell_1_dic_learning} performs well. 
\vspace*{-2mm}

\section{Proof of the main result}
\label{proof_architecture_main_result}

The proof of Theorem \ref{thm_main_result} is based on reducing the minimax estimation problem \eqref{equ_def_minimax_problem} to a specific multiple hypothesis testing problem. 
In particular, we assume that the generating dictionary $\mathbf{D}$ in \eqref{equ_linear_model} is taken from a finite subset $\mathcal{D}_{0} \triangleq \{ \mathbf{D}^{(\dicidx)} \}_{\dicidx \in [L]} \subseteq \mathcal{X}(\mathbf{D}_{0},r)$ for some $L \in \mathbb{N}$. 
This subset $\mathcal{D}_{0}$ is constructed such that (i) any two distinct dictionaries $\mathbf{D}^{(\dicidx)},\mathbf{D}^{(\dicidx')} \in \mathcal{D}_{0}$ are separated 
by at least $\sqrt{8 \minimaxrisk}$, i.e., $\| \mathbf{D}^{(\dicidx)} - \mathbf{D}^{(\dicidx')} \|_{\text{F}} \geq \sqrt{8 \minimaxrisk}$ and (ii) it is hard to detect the generating dictionary $\mathbf{D}$ if it is drawn u.a.r. from $\mathcal{D}_{0}$. 
However, we do not specify a deterministic scheme to construct such a set $\mathcal{D}_{0}$. 
We merely use a probabilistic method to show that there must exist at least one such set $\mathcal{D}_{0}$. 
The existence of $\mathcal{D}_{0}$ then yields, via Lemma \ref{lem_existence_dicationary_set_desiredata}, a relation between the sample-size $\samplesize$ and the remaining model parameters $\measlen$, $\coefflen$, $s$, $\sigma_{a}$, 
$\sigma$ which has to be satisfied such that an estimator with worst-case MSE not exceeding $\minimaxrisk$ may exist.
\begin{figure}[t]
\vspace{-1mm}
\centering
\psfrag{SNR}[c][c][.9]{\uput{3.4mm}[270]{0}{\hspace{0mm}SNR [dB]}}
\psfrag{radius}[c][c][.9]{\hspace*{-1mm}$\sqrt{2 \varepsilon}$}
\psfrag{distance}[c][c][.9]{\hspace*{-1mm}$\sqrt{8 \varepsilon}$}
\psfrag{DicSet}[c][c][.9]{\hspace*{-1mm}$\mathcal{D}$}
\psfrag{D1}[c][c][.9]{\hspace*{2mm}$\mathbf{D}^{(1)}$}
\psfrag{D2}[c][c][.9]{\hspace*{2mm}$\mathbf{D}^{(2)}$}
\psfrag{D3}[c][c][.9]{\hspace*{2mm}$\mathbf{D}^{(3)}$}
\psfrag{D4}[c][c][.9]{\hspace*{2mm}$\mathbf{D}^{(4)}$}
\psfrag{hatD}[c][c][.9]{\hspace*{4mm}$\widehat{\mathbf{D}}(\mathbf{Y})$}
\psfrag{x_m20}[c][c][.9]{\uput{0.3mm}[270]{0}{$-20$}}
\psfrag{x_m_10}[c][c][.9]{\uput{0.3mm}[270]{0}{$-10$}}
\centering
\hspace*{-0mm}\includegraphics[height=6cm]{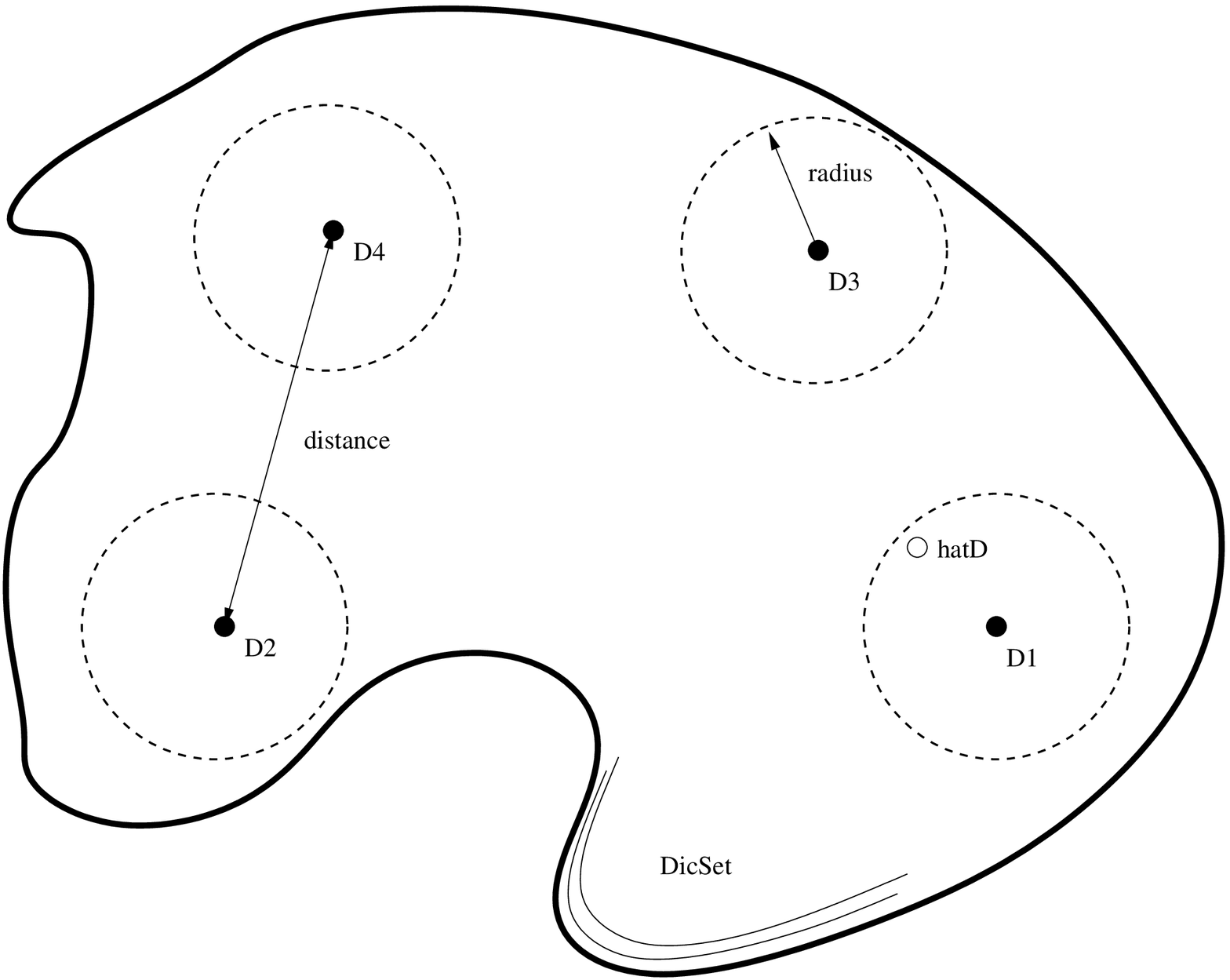}
\vspace*{-4mm}
\renewcommand{\baselinestretch}{1.2}\small\normalsize
  \caption{Finite ensemble $\mathcal{D}_{0}$ of size $L=4$.} 
\label{fig_info_theor_proof_sketch}
\vspace*{-3.5mm}
\end{figure}

In Fig. \ref{fig_info_theor_proof_sketch}, we sketch the idea of this method for the particular case of a subset $\mathcal{D}_{0} \defeq \{ \mathbf{D}^{(1)},\ldots,\mathbf{D}^{(4)} \}$ containing four dictionaries $\mathbf{D}^{(\dicidx)} \in \mathcal{X}(\mathbf{D}_{0},r)$. 
We also show a realization of the estimator $\hat{\mathbf{D}}(\mathbf{Y})$. Two different dictionaries in $\mathcal{D}_{0}$ are separated by at least $\sqrt{8 \varepsilon}$. 
In particular, if $\widehat{\mathbf{D}}(\mathbf{Y})$ is a learning scheme achieving the minimax risk in \eqref{equ_def_minimax_problem}, then the 
minimum distance detector 
\vspace*{-3mm}
\begin{equation}
\argmin_{\mathbf{D}' \in \mathcal{D}_{0}} \| \widehat{\mathbf{D}}(\mathbf{Y}) - \mathbf{D}' \|_{\text{F}} \nonumber
\vspace*{-2mm}
\end{equation} 
recovers the correct dictionary $\mathbf{D} \in \mathcal{D}_{0}$ if $\widehat{\mathbf{D}}(\mathbf{Y})$ belongs to the ball $\mathcal{B}(\mathbf{D},\sqrt{2 \varepsilon})$ (indicated by a dashed circle in Fig. \ref{fig_info_theor_proof_sketch}) centered at $\mathbf{D}$ and with radius $\sqrt{2 \varepsilon}$. 
The information-theoretic method \cite{Wain2009TIT,WangWain2010,BinYuFestschrift1997} of lower bounding the minimax risk $\varepsilon$ consists then in relating the probability $\prob \big\{ \widehat{\mathbf{D}}(\mathbf{Y}) \notin \mathcal{B}(\mathbf{D},\sqrt{2 \varepsilon}) \big\}$ to the mutual information between the observation $\mathbf{Y}$ and the dictionary $\mathbf{D}$ which is assumed to be drawn u.a.r. from $\mathcal{D}_{0}$. 

In particular, our analysis is based on the construction of a finite set $\mathcal{D}_{0} \triangleq \{\mathbf{D}^{(1)}, \ldots,\mathbf{D}^{(L)} \}  \subseteq \mathcal{X}(\mathbf{D}_{0},r)$ of $L$ distinct dictionaries belonging to 
$\mathcal{D}$ (cf. \eqref{equ_def_dictionary_set_normalized_cols}) having the following desiderata: 
\vspace*{-2mm}
\begin{itemize} 
\item For any two dictionaries $\mathbf{D}^{(\dicidx)}, \mathbf{D}^{(\dicidx')} \in \mathcal{D}_{0}$,
\vspace*{-2mm}
\begin{equation} 
\label{equ_cond_frob_norm_diff_delta}
\| \mathbf{D}^{(\dicidx)} - \mathbf{D}^{(\dicidx')} \|^{2}_{\text{F}} \geq \bar{\delta}_{\dicidx,\dicidx'} 8 \varepsilon.
\vspace*{-1mm}
\end{equation} 
\item If the generating dictionary in \eqref{equ_linear_model} is chosen as $\mathbf{D} = \mathbf{D}^{(\dicidx)} \in \mathcal{D}_{0}$, where $\dicidx$ is selected u.a.r. from $[L]$, 
then the conditional mutual information between $\mathbf{Y}$ and $\dicidx$, given $\mathbf{i}$, is bounded as 
\vspace*{-3mm}
\begin{align}
\label{equ_desirata_av_mutual_info_y_index}
\condmiyandlgiveni \leq \eta 
\vspace*{-6mm}
\end{align}
with some given small $\eta$.
\end{itemize}

The following result gives precise conditions on the cardinality $L$ and threshold $\eta$ such that at least one subset $\mathcal{D}_{0} \subseteq \mathcal{X}(\mathbf{D}_{0},r)$ of size $L$ 
satisfying \eqref{equ_cond_frob_norm_diff_delta} as well as \eqref{equ_desirata_av_mutual_info_y_index} is guaranteed to exist.
\begin{lemma}
\label{lem_existence_dicationary_set_desiredata}
Consider a dictionary learning problem based on \eqref{equ_def_par_set_small_perturb} with some $r \leq 1/\sqrt{\coefflen}$. 
Then, for any $\varepsilon$ such that 
\vspace*{-2mm}
\begin{equation} 
\label{equ_cond_epsilon_exist_dic_set}
\varepsilon < r^{2} / 16,\nonumber
\vspace*{-2mm}
\end{equation}
there exists a set $\mathcal{D}_{0} \subseteq \mathcal{X}(\mathbf{D}_{0},r)$ of cardinality $L = e^{\coefflen/32}$ such that 
\eqref{equ_cond_frob_norm_diff_delta} and \eqref{equ_desirata_av_mutual_info_y_index} are satisfied with
$\eta\!=\!32 \varepsilon \samplesize \sparsity \emph{SNR} /\coefflen$.
\end{lemma}
The next result, which is the central argument of the information-theoretic method for lower bounding minimax risk, relates the cardinality $L$ of a subset $\mathcal{D}_{0} \subseteq \mathcal{D}$ to the conditional mutual information $\condmiyandlgiveni$ between 
the observation $\mathbf{Y}$ and a random index $\dicidx$ selecting the generating dictionary u.a.r. from $\mathcal{D}_{0}$.
\begin{lemma}
\label{lem_inequality_Q_log_L}
Consider the dictionary learning problem \eqref{equ_linear_model} with minimax risk $\minimaxrisk$ (\eqref{equ_def_minimax_problem}) and a finite set 
$\mathcal{D}_{0} \subseteq \mathcal{X}(\mathbf{D}_{0},r)$ 
consisting of $L$ distinct dictionaries $\mathbf{D}^{(\dicidx)} \in \mathbb{R}^{\measlen \times \coefflen}$ such that 
\vspace*{-2mm}
\begin{equation}
\label{equ_cond_Frob_norm_squared_greater_8}
\| \mathbf{D}^{(\dicidx)} - \mathbf{D}^{(\dicidx')} \|^{2}_{\text{F}} \geq 8 \bar{\delta}_{\dicidx,\dicidx'} \minimaxrisk. \nonumber
\vspace*{-2mm}
\end{equation}
Then, it holds $\condmiyandlgiveni \geq (1/2) \log_{2}(L) -1.$
\end{lemma}
The proofs of Lemma \ref{lem_existence_dicationary_set_desiredata} and \ref{lem_inequality_Q_log_L} are omitted due to space limitations. 

\emph{Proof of Theorem \ref{thm_main_result}:} According to Lemma \ref{lem_existence_dicationary_set_desiredata}, for any $\varepsilon < r^{2}/8$, with $r \leq 1/\sqrt{p}$, there exists a set $\mathcal{D}_{0} \subseteq \mathcal{X}(\mathbf{D}_{0},r)$ of cardinality $L = e^{\coefflen/32}$ 
satisfying \eqref{equ_cond_frob_norm_diff_delta} and \eqref{equ_desirata_av_mutual_info_y_index} with 
$\eta=32 \samplesize \sparsity \mbox{SNR}^2 \varepsilon/\coefflen$. 
Applying Lemma \ref{lem_inequality_Q_log_L} to the set $\mathcal{D}_{0}$ yields, in turn, 
\vspace*{-3mm}
\begin{equation} 
32\samplesize \sparsity \mbox{SNR} \varepsilon/\coefflen \geq \condmiyandlgiveni \geq (1/2) \log_{2}(L) -1  \nonumber 
\vspace*{-3mm}
\end{equation}
implying 
\vspace*{-3mm}
\begin{equation}
\varepsilon  \geq  \frac{\mbox{SNR}^{-1}}{32 \samplesize \sparsity} \coefflen ((1/2) \log_{2}(L) -1). \nonumber
\vspace*{-3mm}
\end{equation}
Since 
\vspace*{-4mm}
\begin{equation}
(1/2) \log_{2}(L) -1 \geq 0.7 \coefflen/32 - 1 \stackrel{\eqref{equ_cond_coefflen_measlen}}{\geq}  0.2 \coefflen/32, \nonumber
\vspace*{-2mm}
\end{equation} 
we arrive at \eqref{equ_conditions_s_2_first_charac_minimax_risk}. 

\section{Numerical Experiments}

\begin{figure}[t]
\vspace{-1mm}
\centering
\psfrag{xlabel}[c][c][.9]{\uput{3.4mm}[270]{0}{\hspace{0mm}$\log_{2}N$}}
\psfrag{DirHad01}[c][c][.9]{\uput{0mm}[0]{0}{\hspace{-5.6mm}$\mathbf{D} \!=\!\mathbf{D}_{2}\mbox{, SNR}\!=\!20\mbox{dB}$}}
\psfrag{DirHad001}[c][c][.9]{\uput{0mm}[0]{0}{\hspace{-6.4mm}$\mathbf{D} \!=\!\mathbf{D}_{2}\mbox{, SNR}\!=\!40\mbox{dB}$}}
\psfrag{Dir01}[c][c][.9]{\uput{0mm}[0]{0}{\hspace{-3.8mm}$\mathbf{D} \!=\!\mathbf{I}\mbox{, SNR}\!=\!20\mbox{dB}$}}
\psfrag{Dir001}[c][c][.9]{\uput{0mm}[0]{0}{\hspace{-4.6mm}$\mathbf{D}\!=\!\mathbf{I}\mbox{, SNR}\!=\!40\mbox{dB}$}}
\psfrag{ylabel}[c][c][.9]{\uput{3mm}[90]{0}{\hspace{-9mm}$\log_{2}Ê\varepsilon$}}
\psfrag{0}[c][c][.9]{\uput{0.3mm}[180]{0}{$0$}}
\psfrag{-1}[c][c][.9]{\uput{0.3mm}[180]{0}{$-1$}}
\psfrag{-2}[c][c][.9]{\uput{0.3mm}[180]{0}{$-2$}}
\psfrag{-3}[c][c][.9]{\uput{0.3mm}[180]{0}{$-3$}}
\psfrag{-4}[c][c][.9]{\uput{0.3mm}[180]{0}{$-4$}}
\psfrag{-5}[c][c][.9]{\uput{0.3mm}[180]{0}{$-5$}}
\psfrag{-6}[c][c][.9]{\uput{0.3mm}[180]{0}{$-6$}}
\psfrag{-7}[c][c][.9]{\uput{0.3mm}[180]{0}{$-7$}}
\psfrag{-8}[c][c][.9]{\uput{0.3mm}[180]{0}{$-8$}}
\psfrag{-9}[c][c][.9]{\uput{0.3mm}[180]{0}{$-9$}}
\psfrag{-10}[c][c][.9]{\uput{0.3mm}[180]{0}{$-10$}}
\psfrag{7}[c][c][.9]{\uput{0.3mm}[270]{0}{$7$}}
\psfrag{8}[c][c][.9]{\uput{0.3mm}[270]{0}{$8$}}
\psfrag{9}[c][c][.9]{\uput{0.3mm}[270]{0}{$9$}}
\psfrag{10}[c][c][.9]{\uput{0.3mm}[270]{0}{$10$}}
\psfrag{11}[c][c][.9]{\uput{0.3mm}[270]{0}{$11$}}
\psfrag{12}[c][c][.9]{\uput{0.3mm}[270]{0}{$12$}}
\psfrag{13}[c][c][.9]{\uput{0.3mm}[270]{0}{$13$}}
\psfrag{14}[c][c][.9]{\uput{0.3mm}[270]{0}{$14$}}
\psfrag{x_m_10}[c][c][.9]{\uput{0.3mm}[270]{0}{$-10$}}
\centering
\hspace*{-6mm}\includegraphics[height=7cm,width=10cm]{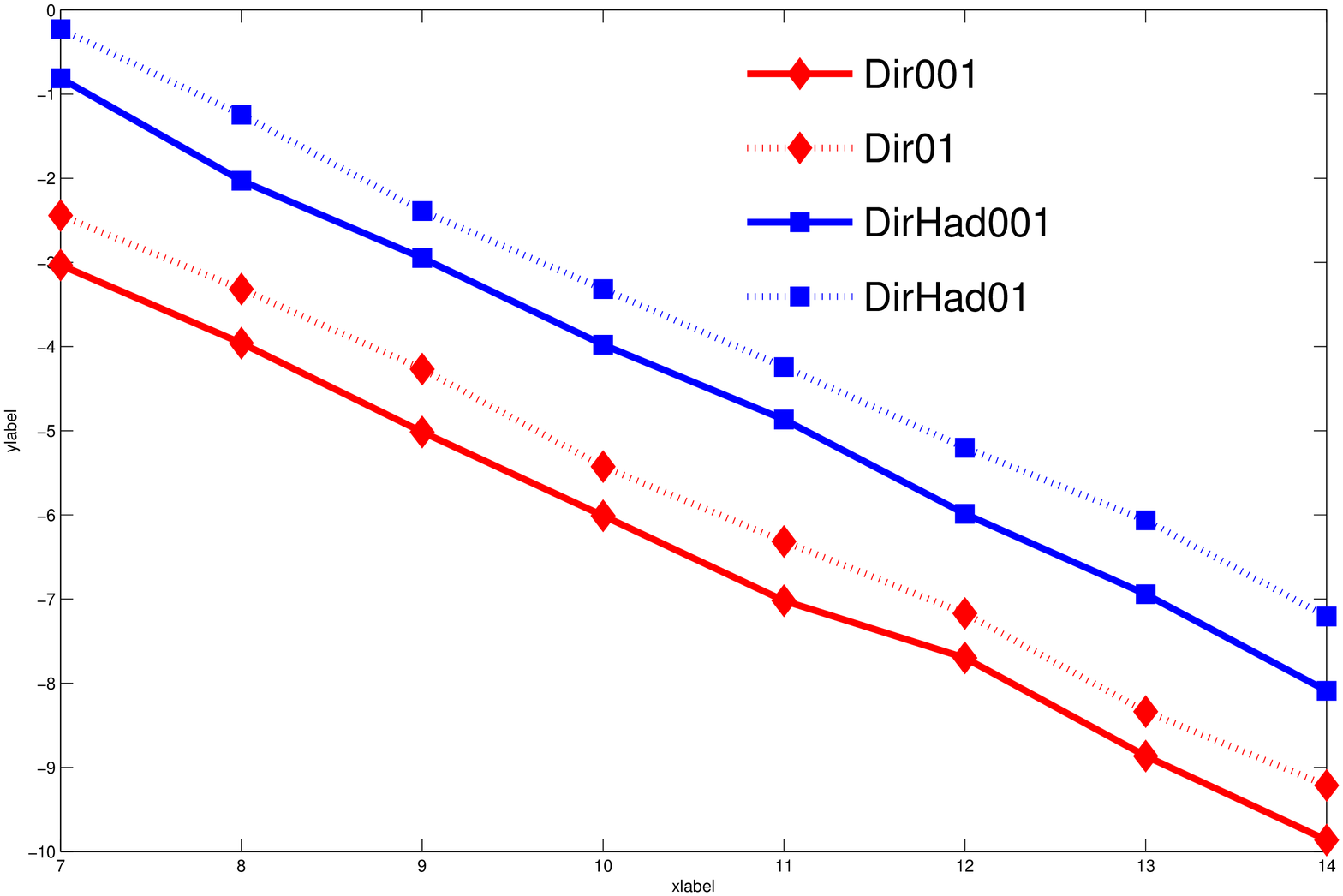}
\vspace*{-8mm}
\renewcommand{\baselinestretch}{1.2}\small\normalsize
  \caption{MSE curves of the ITKM learning scheme for $\measlen=8$.} 
\label{fig_iktm_MSE}
\vspace*{-8mm}
\end{figure}

One of the uses of the lower bound on the minimax risk stated in Theorem \ref{thm_main_result} is that it allows for an assessment of the performance 
of practical learning schemes. In this section we compare the lower bound \eqref{equ_conditions_s_2_first_charac_minimax_risk} with the actual MSE of an (locally) efficient learning scheme $\widehat{\mathbf{D}}_{\text{ITKM}}(\mathbf{Y})$, termed \emph{iterative thresholding and K-means} (ITKM) algorithm, which has been proposed recently \cite{SchnassIKTM2014}. 
We applied the ITKM algorithm with sparsity parameter $\tilde{s}=1$, using oracle initialization and signal normalization\footnote{For background and notation, we refer to \cite{SchnassIKTM2014}.}, to a data matrix $\mathbf{Y} \in \mathbb{R}^{\measlen \times \samplesize}$, with $\measlen=8$, whose columns are independent realizations of $\mathbf{y}$ according to \eqref{equ_linear_model} with $\sparsity=2$. For the underlying generating dictionary $\mathbf{D}$ we choose the identity matrix $\mathbf{I}$ and, in a second experiment, the concatenation of the identity matrix and the $\measlen \times \measlen$ normalized Hadamard matrix $\mathbf{F}_{\measlen}$\footnote{For $\measlen$ being a power of $2$, the Hadamard matrix $\mathbf{F}_{\measlen}$ is defined recursively by $\mathbf{F}_{1} = \begin{pmatrix} 1 & 1 \\ 1 & -1 \end{pmatrix}$ and $\mathbf{F}_{\measlen} =  \mathbf{F}_{1} \otimes  \mathbf{F}_{\measlen/2}$.}, i.e., $\mathbf{D} \!=\! \mathbf{D}_{2} \! \defeq \! \big[ \mathbf{I} \,\, \sqrt{1/\measlen} \mathbf{F}_{\measlen}\big]$. For both choices for the generating dictionary we set $\measlen \!= \!8$ and $\sparsity\!=\!2$. In Fig.\ \ref{fig_iktm_MSE}, we plot the actual MSE $\varepsilon(\mathbf{D},\widehat{\mathbf{D}}_{\text{ITKM}}(\cdot))$ for varying sample-size $N$ and different values of the SNR. The bound \eqref{equ_conditions_s_2_first_charac_minimax_risk} correctly predicts the slope $1/N$ of the curves. However, the absolute position of the lower bound \eqref{equ_conditions_s_2_first_charac_minimax_risk} is significantly below that of the actual MSE curves. While this could mean that the performance of ITKM is far from optimum, there is also the possibility that the lower bound \eqref{equ_conditions_s_2_first_charac_minimax_risk} can be tightened (made higher) considerably by 
taking also the term $I( \dicidx ; \mathbf{\suppidx} | \mathbf{Y})$ in \eqref{equ_decomp_mutual_inf_chain_rule} into account. 

\vspace*{-4mm}
\section{Conclusion}
\vspace*{-3mm}

We derived a lower bound on the minimax risk for dictionary learning, which seems to be the first result of this kind. 
This lower bound yields, in turn, a characterization of the required sample-size, i.e., the sample-complexity, such that accurate learning schemes, regardless of 
computational complexity, may exist. Comparing our results with the sample-complexity of some popular learning schemes, which are mainly based on minimizing \eqref{equ_minimization_ell_1_dic_learning}, reveals that there may be other algorithms requiring significantly fewer observations. Finally, we note that our lower bound 
complements the sufficient conditions on the sample-complexity for dictionary learning derived in \cite{VainManBruck11}.
\vspace{-3mm}
\section{Acknowledgment}
\vspace{-2mm}

The authors would like to thank Karin Schnass for sharing here expertise on practical dictionary learning schemes and for providing some simulation 
results. 

\renewcommand{\baselinestretch}{0.9}\normalsize\footnotesize
\bibliographystyle{IEEEbib}
\bibliography{/Users/ajung/Arbeit/LitAJ_ITC.bib,/Users/ajung/Arbeit/tf-zentral}

\end{document}